\documentclass[10pt,letterpaper]{article}
\usepackage[top=0.85in,left=2.75in,footskip=0.75in]{geometry}
\usepackage[document]{ragged2e}
\usepackage{soul}
\usepackage{caption}
\usepackage{upgreek}

\usepackage{amsmath,amssymb}

\usepackage{changepage}

\usepackage[utf8x]{inputenc}

\usepackage{textcomp,marvosym}

\usepackage{cite}

\usepackage{nameref,hyperref}

\usepackage[right]{lineno}

\usepackage{microtype}
\DisableLigatures[f]{encoding = *, family = * }

\usepackage[table]{xcolor}

\usepackage{array}

\newcolumntype{+}{!{\vrule width 2pt}}

\newlength\savedwidth

\newcommand\thickhline{\noalign{\global\savedwidth\arrayrulewidth\global\arrayrulewidth 2pt}%
\hline
\noalign{\global\arrayrulewidth\savedwidth}}

\raggedright
\usepackage[aboveskip=1pt,labelfont=bf,labelsep=period,justification=raggedright,singlelinecheck=off]{caption}

\makeatletter
\renewcommand{\@biblabel}[1]{\quad#1.}
\makeatother

\usepackage{comment}

\def\BibTeX{{\rm B\kern-.05em{\sc i\kern-.025em b}\kern-.08em
    T\kern-.1667em\lower.7ex\hbox{E}\kern-.125emX}}

\newcommand{\Rmnum}[1]{\expandafter\@slowromancap\romannumeral #1@}

\usepackage{lastpage,fancyhdr,graphicx}
\usepackage{epstopdf}
\fancyheadoffset[L]{2.25in}
\fancyfootoffset[L]{2.25in}
\begin{document}
\vspace*{0.2in}

\begin{flushleft}
{\Large
\textbf\newline{ A laser-microfabricated electrohydrodynamic thruster for centimeter-scale aerial robots }
}
\newline

Hari Krishna Hari Prasad\textsuperscript{1\Yinyang}
Ravi Sankar Vaddi\textsuperscript{2\Yinyang},
Yogesh M Chukewad\textsuperscript{1\Yinyang},
Elma Dedic\textsuperscript{1},
Igor Novosselov\textsuperscript{2}, and
Sawyer B Fuller\textsuperscript{1*}
\\
\bigskip
\textbf{1} Autonomous Insect Robotics (AIR) Lab, Department of Mechanical Engineering, University of Washington, Seattle, WA 98195, USA
\\
\textbf{2} Novosselov Research Group, Department of Mechanical Engineering, University of Washington, Seattle, WA 98195, USA
\bigskip
\\
\Yinyang These authors contributed equally to this work.
\\
* minster@uw.edu

\end{flushleft}

\justify
\section*{Abstract}
To date, insect scale robots capable of controlled flight have used flapping wings for generating lift, but this requires a complex and failure-prone mechanism. 
A simpler alternative is electrohydrodynamic (EHD) thrust, which requires no moving mechanical parts. In EHD, corona discharge generates a flow of ions in an electric field between two electrodes; the high-velocity ions transfer their kinetic energy to neutral air molecules through collisions, accelerating the gas and creating thrust. We introduce a fabrication process for EHD thruster based on 355~nm laser micromachining and our approach allows for greater flexibility in materials selection. Our four-thruster device measures 1.8 $\times$ 2.5~cm and is composed of steel emitters and a lightweight carbon fiber mesh. The current and thrust characteristics of each individual thruster of the quad thruster is determined and agrees with Townsend relation. The mass of the quad thruster is 37~mg and the measured thrust is greater than its weight (362.6~$\upmu \text{N}$). The robot is able to lift off at a voltage of 4.6 kV with a thrust to weight ratio of 1.38.

\section*{Author summary}

Agile small aerial robots can be useful in exploring areas where humans cannot reach. Applications of these robots include gas leak detection in pipes, search and rescue in case of natural disasters, etc. There have been recent developments in flapping wing robots at an insect scale, but these are mechanically complex and difficult to fabricate. Recently, researchers have reported progress in developing aerial vehicles using electrohydrodynamic (EHD) thrust. This phenomenon, also known as ionic wind, produces ions that collide with neutral air molecules to create thrust. As there are no mechanically moving parts, this propulsion scheme eliminates fatigue failure, simplifies assembly, and makes the robots more robust to crashes. Here we report a new means to fabricate these devices more rapidly using a laser-based process, demonstrating a four-thruster robot able to lift its own weight.

\section*{Introduction}
Insect-scale robotics has been an area of interest for its possible uses in agriculture, search and rescue, and biomedicine, among other areas. The small size and reduced manufacturing cost of insect robots have facilitated microrobotic research. To date, the primary emphasis in insect-scale robotics has been on drawing inspiration from biology, because biology has found solutions whose existence proves they work. One example is a flapping-wing robot fly\cite{ma2013controlled}. Robots of this type have subsequently incorporated onboard sensors for flight stabilization \cite{oceli}, and lifted off for the first time without needing a wire tether reaching to the ground using a laser power source \cite{james2018liftoff}, and subsequently, incandescent light~\cite{Jafferis2019}. Other developments include using explosives to break the surface tension of water \cite{chen2017biologically} and RoboFly, which is capable of performing multi-modal locomotion including walking in addition to flying \cite{chukewad2018}. While flapping wings are well suited to insect-sized aerial vehicles, they impose a significant cost in terms of mechanical complexity \cite{WoodPico2012}. In this paper, we focus on an alternative means of generating thrust that is not seen in biology: electrohydrodynamic (EHD) thrust. EHD thrust requires sustained high voltage, which may be why it is not observed in biology. From an engineering perspective, EHD has the appealing characteristic that it requires no moving mechanical parts, simplifying fabrication. A recent advance indicating the promise of EHD thrust was fully EHD-powered 2.5~kg airplane~\cite{MIT2018}

The work in~\cite{drew2018} reported the demonstration of a very small EHD-based aircraft, the ``Ionocraft" measuring only 3~cm across and consisting of four-thrusters. It was able to take off using an external power source. The strength of that work was how the low outflow velocity from low-voltage EHD thrusters was a good match to the low mass of that device. Additionally, EHD is potentially simpler than flapping-wing flight because even a four-thruster device consists of only a single moving part. However, the device in~\cite{drew2018} was fabricated using expensive semiconductor-based cleanroom fabrication facilities. Significant engineering development is required before a small EHD-powered robot can perform aerial locomotion without wires and fully autonomously, which will be necessary for such robots to have a useful application. More rapid and less expensive methods to build the robots with EHD propulsion could facilitate faster design iteration times, which is highly desired when technology is still in its infancy.

\begin{figure}[!htbp]
\centering
\includegraphics[width=13cm]{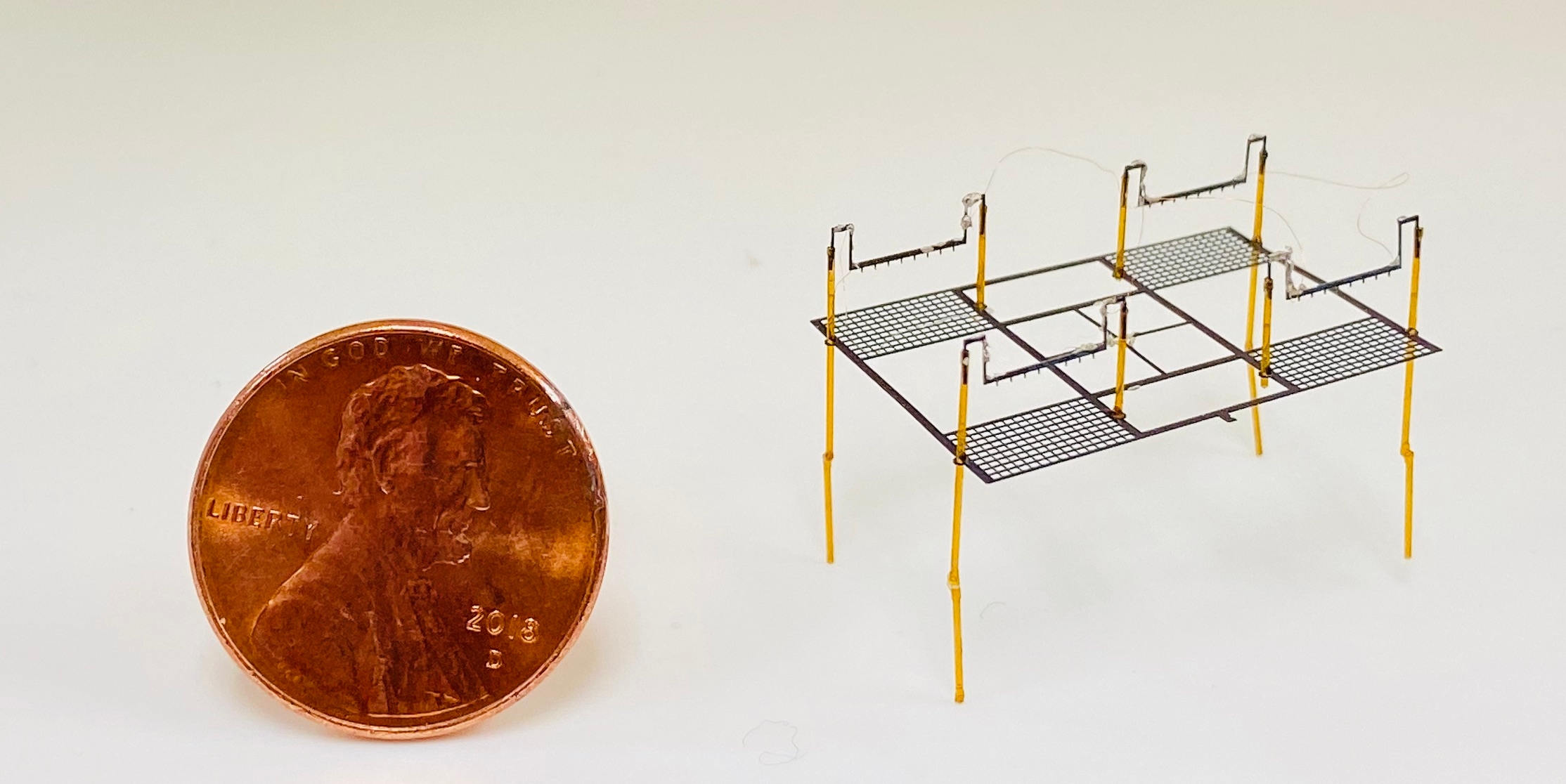}
\caption{An assembled Quad-thruster robot next to a U.S.penny
The $1.8\times 2.5$~cm quad-thruster having a mass of 37~mg is shown. The robot components consist of a carbon fiber collector grid, four blue tempered steel emitters, and eight fiber optic glass tubes. All components are hand-assembled using external jigs.}
\label{fig1}
\end{figure}

In this paper, we utilize laser machining fabrication to build centimeter-sized aerial robots with EHD thrusters. The process allows for use of a greater variety of electrode materials and eliminates the need for a cleanroom facility. For example, it allows fabricating a complete four-thruster device in a matter of minutes. The quad-thruster robot presented in this paper, as shown in Fig.~\ref{fig1} has generated thrust exceeding its weight. We report the fabrication methodology and the experimental measurements of the corona current, thrust and energy transfer efficiency for each individual thruster of the quad-thruster robot. Finally, we present the takeoff of our quad-thruster robot.

\section*{Electrohydrodynamics (EHD)}

Electrohydrodynamics (EHD) is an interdisciplinary field describing the interaction of fluids with an electric field. Insights into complex multi-physics interactions are essential for understanding EHD flows: (1) ion generation; (2) the ion motion in the electric field; (3) the interaction between the motion of ions and the neutral molecules; and (4) the inertial and viscous forces in the complex flow.

\subsection*{Corona discharge driven flow}
 
Corona discharge generates a flow of ions in a strong electric field between two electrodes; the high-velocity ions transfer their kinetic energy to the neutral air molecules by collisions that accelerate the gas in the direction of ion drift. This electrohydrodynamic (EHD) flow propulsion phenomenon, also referred to in the literature as ionic wind, is used in many practical applications, such as convective cooling \cite{go2008enhancement,jewell2008cfd}, electrostatic precipitators (ESP) \cite{vaddi2019particle}, plasma-assisted combustion \cite{ju2015plasma}, airflow control \cite{moreau2007airflow, roth2003aerodynamic}, and as a turbulent boundary layer actuators \cite{choi2011turbulent}. The corona induced EHD flow converts electric energy into kinetic energy directly and requires no moving parts. The voltage-current relation during the corona discharge characterizes the ion motion between the electrodes. This phenomenon has been studied since the early 20th century. The classic relationship was derived by Townsend \cite{townsend1915electricity} in 1914 and validated for a coaxial corona configuration. Some recent studies modify Townsend’s quadratic relationship to better describe the   relationship for different electrode configurations \cite{yanallah2012semi,zhang2015characteristics,martins2012simulation,guan2018analytical}.  A generalized analytical model for voltage to current and voltage to velocity relationship for EHD driven flow has been recently described \cite{guan2018analytical}; the analytical model has a good agreement with the experimental data in the accelerating flow regions (EHD dominated flow). Previous studies have reported that maximum velocity for point-to-ring electrode configuration was recorded at ~9 m/s \cite{guan2018experimental} and have assessed the use of ionic winds in propulsion applications \cite{moreau2013electrohydrodynamic}.

Stuetzer \cite{stuetzer1959ion} presented the first experimental and theoretical analysis of pressure drag produced by the ions, where he determined the pressure generation over a wide range of carrier media. Previous work performed by Masuyama \cite{masuyama2013performance} determined the achievable thrust to power ratios of EHD propulsion on the orders of 5-10 N.kW$^{-1}$. Thrust to power ratio was found to be dependent on electrode distance and the potential difference between the electrodes. Similar results were observed for an ionocraft with a wireless power supply onboard  and transmitted power up to 100 W to ionocraft at the voltages up to 12 kV \cite{khomich2018atmosphere}. The EHD propulsion can be utilized for UAV propulsion; the experimentally measured maximum thrust density of 15 N.m$^{-3}$ was reported recently \cite{gilmore2015electrohydrodynamic}. Drew et al. showed that higher thrust density can be achieved for insect-scale robots \cite{drew2017first,drew2018} and EHD flow can be used for flight control. 

\subsection*{Electrohydrodynamic Force}

A one dimensional model for an EHD thruster yields an expression in terms of the current, distance between the anode and cathode. Space charge effect is ignored here, however, it can be important at high electric field strengths. The current is determined by integrating charge density
\begin{equation}
I = \int {J.dA} = \int {\rho_e E \mu dA}
\label{current_eq}
\end{equation}
where $\rho_e$\ is the charge density, $\mu$ is the ion mobility in the air, $E$ is the electric field. Ion mobility value of $\mu = 2\times10^{-4}$~m$^2$/V-s is used here. For  energy transfer efficiency analysis, consider that thrust is equal to the Coulomb force acting on the volume of  fluid between the anode and cathode

\begin{equation}
F = \int {\rho_e E dV}= \frac{Id}{\mu}
\label{force_eq}
\end{equation}
where $F$ is the thrust, $I$ is the ion current, $d$ is the distance between electrodes. The corona power can be written as in Eq. \ref{power_eqn}, and efficiency which is defined as $F/P$ is given by equation Eq. \ref{forceoverP_eq}.

\begin{equation}
P = IV
\label{power_eqn}
\end{equation}
\begin{equation}
\frac{F}{P}=\frac{d}{\mu V}=\frac{1}{E \mu} 
\label{forceoverP_eq}
\end{equation}
where \textit{E} is the electric field strength, and \textit{V} is the applied voltage. Drew et al. report the minimum efficiency in their design should be 2~mN/W. The analysis sheds insight into the importance of the electrode distance and applied voltage. Eq. \ref{forceoverP_eq} shows that for the larger electrode spacing higher efficiency values can be reached as observed by Guan et al.  \cite{guan2018experimental}. Related to electrode configuration, it is important to revisit Townsend's relations \cite{townsend1914xi} 

\begin{equation}
I=CV(V-V_{crit})
\label{current_eq_townsend}
\end{equation}
where $V_{crit}$ is the onset voltage and \textit{V} is the voltage applied. \textit{C} is a constant related to the geometry of the electrodes \cite{masuyama2013performance}.  Thrust can be determined using Townsend's relation
\begin{equation}
F=\frac{CV(V-V_{crit})d}{\mu}
\label{force_eq_townsend}
\end{equation}
 In practical thruster design to achieve maximum thrust, the constant $C$  needs to be maximized and $V_{crit}$ needs to be minimized. Among the considerations related to the thruster design are the effects of non-linear ionization region, secondary flow structures, cathode blockage ratio, the transition from glow to streaming corona discharge and eventually to sparkover. The full optimization of the EHD thruster is beyond the scope of this paper. 

\section*{Fabrication}
Our fabrication process emphasizes speed and simplicity by minimizing the number of components and fabrication time.

Previous work has used a silicon-on-insulator process for the fabrication of EHD thruster \cite{drew2017first}. The emitter and collector electrodes were made from silicon patterned with a photolithographic mask. After a deep reactive ion etch (DRIE) process to ablate through the wafer, the electrodes are once more etched with hydrofluoric acid (HF). In that work, insulating standoffs to separate the emitter from the collector were made from fused silica capillary tubing with an outer diameter of 400~$\upmu \text{m}$. Connections between the tubing and electrodes were made with UV-curable epoxy. Power connections are made with the application of silver epoxy. An external jig was used to align the assembly. Our robot has 13 components (8 capillary tubes, 4 emitters and 1 collector) in comparison to the most recently reported design for the ``Ionocraft" that has 41 components (including sensor components) \cite{drew2018}.

\begin{figure}[!htbp]
\centering
\includegraphics[width=13cm]{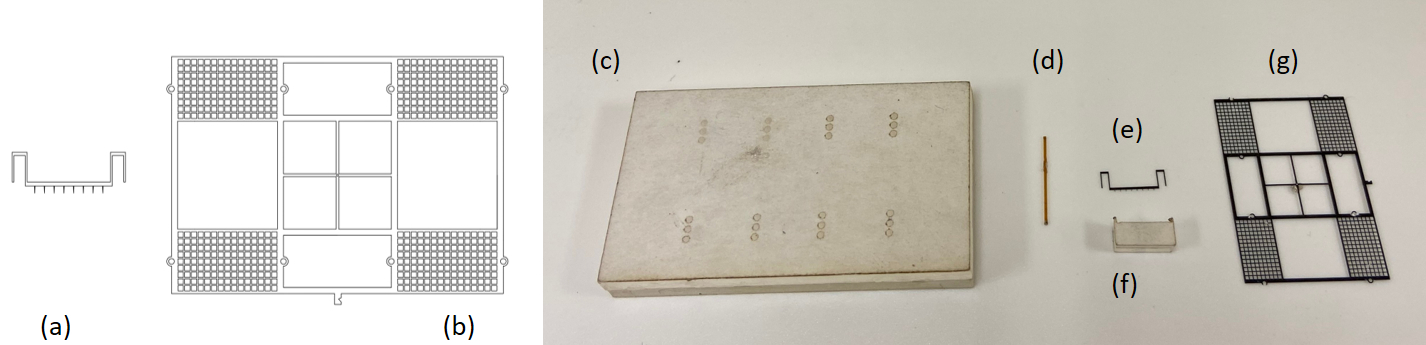}
\caption{\textbf{Individual components of the quad-thruster robot.}
Top view of (a) an emitter drawing, (b) a quad-thruster collector grid drawing. (c) A holding jig used for placing eight glass fiber-optic poles shown in (d) and also for keeping the grid in a plane perpendicular to the poles. (d) One of the eight poles required for the assembly. (e) One of the four emitters involved in the robot. (f) A spacing jig used for keeping emitters at a uniform distance from the grid. (g) A quad-thruster collector grid.}
\label{fig2}
\end{figure}

Here, we use laser micro-machining instead of lithographically-patterned silicon. Our laser is a diode-pumped solid-state (DPSS) frequency tripled Nd:Yag laser with 355~nm wavelength (PhotoMachining, Inc., Massachusetts). The DPSS laser output power is 2~W, its beam diameter is  20~$\upmu \text{m}$, and position repeatability is about 3~$\upmu \text{m}$. With this system, we are able to machine both the emitters and collectors in about ten minutes. Our proposed methodology involves machining using the following steps:

\begin{enumerate}
    \item Blue tempered steel shim and carbon fiber sheet are laser machined with features for emitters and grid respectively, using the DPSS laser.  Their corresponding CAD drawings and actual machined parts are shown in Fig. \ref{fig2} (a), (b) and Fig. \ref{fig2} (e), (g), respectively
    
    \item Jig-1 (holding jig) and jig-2 (spacing jig) (Fig.~\ref{fig2}~(c) and Fig.~\ref{fig2}~(f), respectively) are fabricated out of a sheet of acrylic using a standard CO$_2$ laser cutter.
    
    \item Poles made out of glass fiber optic tubing of an inner diameter of 250~$\upmu \text{m}$ and an outer diameter of 350~$\upmu \text{m}$ are used for maintaining a uniform gap between the electrodes. One of the poles is shown in Fig.~\ref{fig2}~(d).

\end{enumerate}

\section*{Design Analysis}

In this section, we discuss various parameters involved in the design and assembly of the robot. We walk through design considerations to optimize the thrust generated while selecting values for these parameters. 

\subsection*{Emitter}

The emitter (corona electrode) material must be rigid, conductive, and with high curvature features (points). The design analysis of the emitter electrode included material, curvature, number of emitter tips, and orientation. 

The material initially used was a 50~$\upmu \text{m}$ stainless steel but was changed to a 100~$\upmu \text{m}$ blue tempered stainless steel as it was a stiffer and more durable material. We explored different tip angles, starting with 30$^\circ$, before reducing to 10$^\circ$ and then 5$^\circ$. A smaller radius of curvature of an emitter tip creates a stronger electric field gradient and a high ion concentration. Due to the limitations on the laser beam diameter used to fabricate the emitter and local heating due to the beam, we found that 5$^\circ$ was the sharpest tip that the machine could fabricate.

The number of emitter tips corresponded to the number of electric field localization for corona discharge to occur. We explored different numbers of tips. In each case, the thruster showed similar thrust values, so we settled on eight tips. The last factor evaluated was the emitter tip orientation. The emitter design in the baseline iteration had electrode tips that were designed to be in the plane parallel to the collector grid. This case involved ions having an initial velocity component in the horizontal direction. It was concluded that the horizontal velocity component and hence the kinetic energy loss can be avoided by pointing the tips directly towards the grid.

\subsection*{Collector}
The collector is the heaviest of all the components in the robot. It should have a low blockage ratio to allow the thrust-causing air molecules to flow through it while remaining stiff. We explored different grid spacing and material thicknesses. 

We started with a collector grid with 150~$\upmu \text{m}$ spacing between grid marks made out of readily available 50~$\upmu \text{m}$ stainless steel. Due to issues with weight and bending of the stainless steel grid from the strain of other components, we switched to unidirectional carbon fiber reinforced composite. The carbon fiber sheet was made by laying up the 69~GSM (69~g/m$^2$) carbon fiber (TenCate M49J) in 0-90-0 directions. After curing, this lay-up measures about 180~$\upmu \text{m}$ thick. The mass of the single thruster carbon fiber grid with this configuration was 5.9~mg, compared to the previous 8.3~mg stainless steel grid. 
We further optimized spacing and weight by reducing the spacing from 150~$\upmu \text{m}$ to 100~$\upmu \text{m}$. We were unable to achieve a functional collector using thinner, 90~$\upmu \text{m}$ carbon fiber and reduced grid spacing of 50~$\upmu \text{m}$ because of excessive breakage during fabrication (when laser cutter has a spot radius of 20~$\upmu \text{m}$ which ends up removing material unevenly on both sides of a grid line). There are 15$\times$9 square openings in the collector grid each with a side length of 100~$\upmu \text{m}$. This works out to a total flow area of 21.6~mm$^2$ corresponding to a blockage of 32.75$\%$.

\subsection*{Quad-thruster design}
\begin{figure}[!h]
\centerline{\includegraphics[width=8cm]{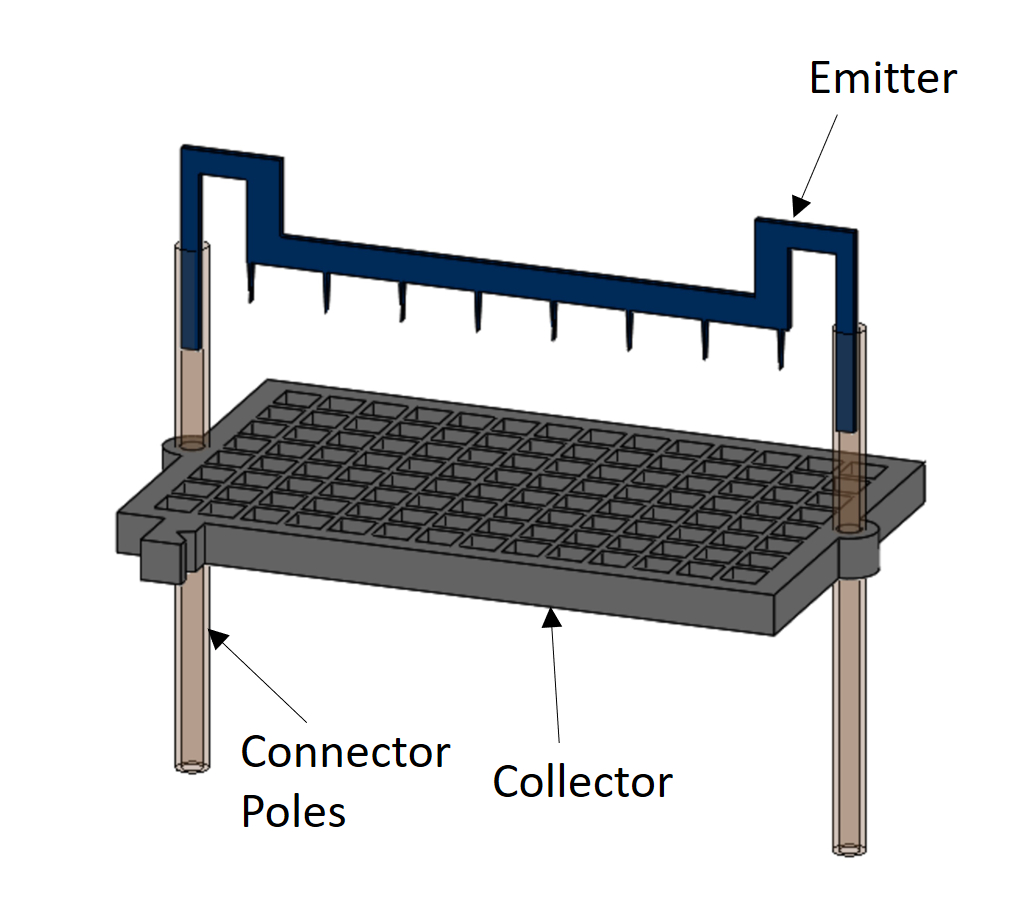}}
\caption{
Dimetric view of collector grid, emitter electrode, and connector poles for the current iteration of a single thruster.}
\label{fig3}
\end{figure}

After individual thrusters were designed, the next design steps involved putting four of the single thrusters together to make a quad-thruster. In the quad thruster, each single thruster was separated by 7~mm and the inter-electrode spacing is 3.5~mm. Inter-electrode spacing is chosen iteratively based on the observed thrust to weight ratio for the current robot design. Effect of inter-thruster spacing (7~mm) and orientation is neither explored nor optimized in this current work. We also created a single thruster, as shown in Fig.~\ref{fig3}, for testing and performance characterization. 

\section*{Assembly} \label{assembly}

In this section, we discuss the assembly process for the quad-thruster. The assembly takes about 15 minutes to complete after the components are fabricated. The steps involved in the assembly process are summarized below.

\begin{figure}[!h]
\centering
\includegraphics[width=13cm]{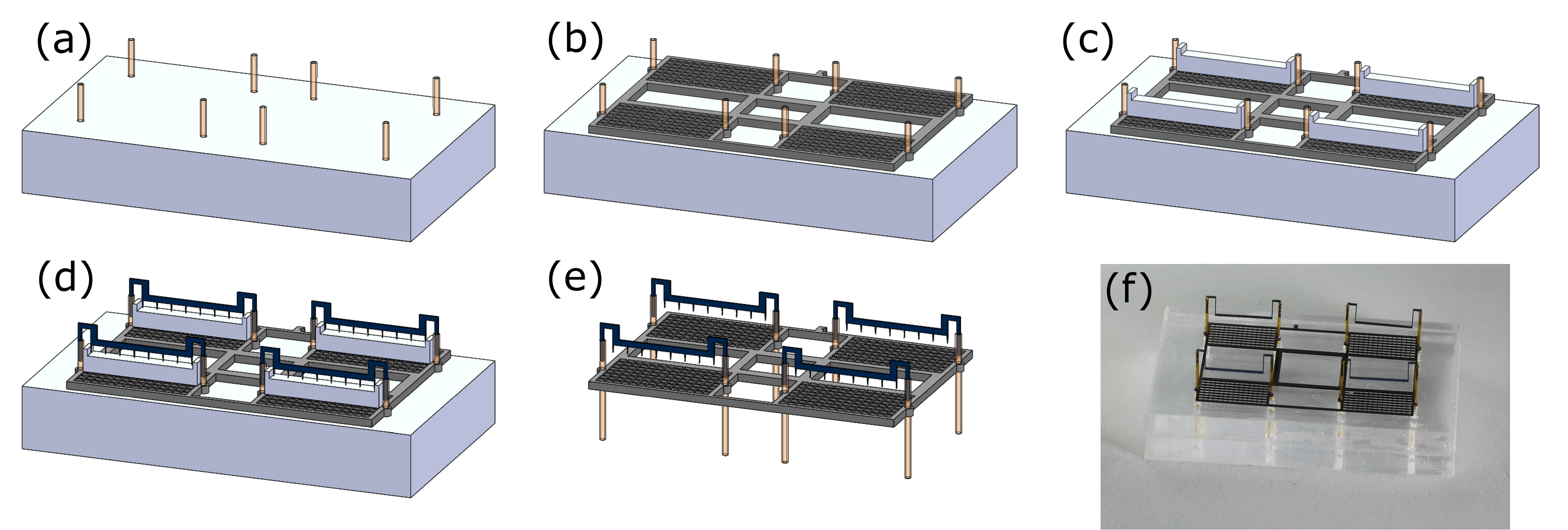}
\caption{ \textbf{Assembly steps of the quad-thruster robot.}
(a) Fiber-optic glass poles placed into the holes of the holding jig. (b) Grid is then placed on holding jig through the poles. Note: grid thickness exaggerated for 3D view. (c) Four spacing jigs placed on each single collector grid. (d) Four emitters are slid into the poles on top of spacing jigs. (e) Dimetric view of a quad-thruster after the jigs are removed. (f) Picture of a quad-thruster fully assembled in the external jig-1 that is used for assembly.}
\label{fig4}
\end{figure}

\begin{enumerate}

    \item The set of four poles are placed through holes on jig-1 as shown in Fig.~\ref{fig4}~(a). 
    
    \item The grid is carefully placed on the holding jig through the poles, it is then glued down with these poles from top to avoid accidentally gluing poles of the grid with the jig. It then looks as shown in Fig.~\ref{fig4}~(b).
    
    \item Four spacing jigs (having a height of 3.5~mm)  are now placed on each single collector grid as shown in Fig.~\ref{fig4}~(c).
    
    \item Four emitters are now slid into the poles on top of single collector grids as shown in Fig.~\ref{fig4}~(d). It is made sure that all of the tips are in contact with their spacing jigs. These emitters are then glued down (Cyanoacrylate) with the poles.
    
    \item Once the glue is dry, all four spacing jigs are slid out, and the whole assembly is then taken out of the holding jig. The assembly looks like the CAD shown in Fig.~\ref{fig4}~(e). Fig.~\ref{fig4}~(f) shows a picture of an actual assembly sitting on a holding jig.
    
    \item The whole system is powered through external tethers. The quad-thruster has 2 external wires; a 58-gauge copper wire is attached to one of the ends of the emitters such that the connection is closer to the center of the collector grid, and another 58-gauge copper wire is attached to the center of the collector grid. Each wire is placed over electrode and silver paste is added using a probe tip. After an electrical connection is established, a bit of glue is added to further reinforce the joint.
\end{enumerate}

\section*{Experimental Method and Results} 
\begin{figure}[!htbp]
\centerline{\includegraphics[width=10cm]{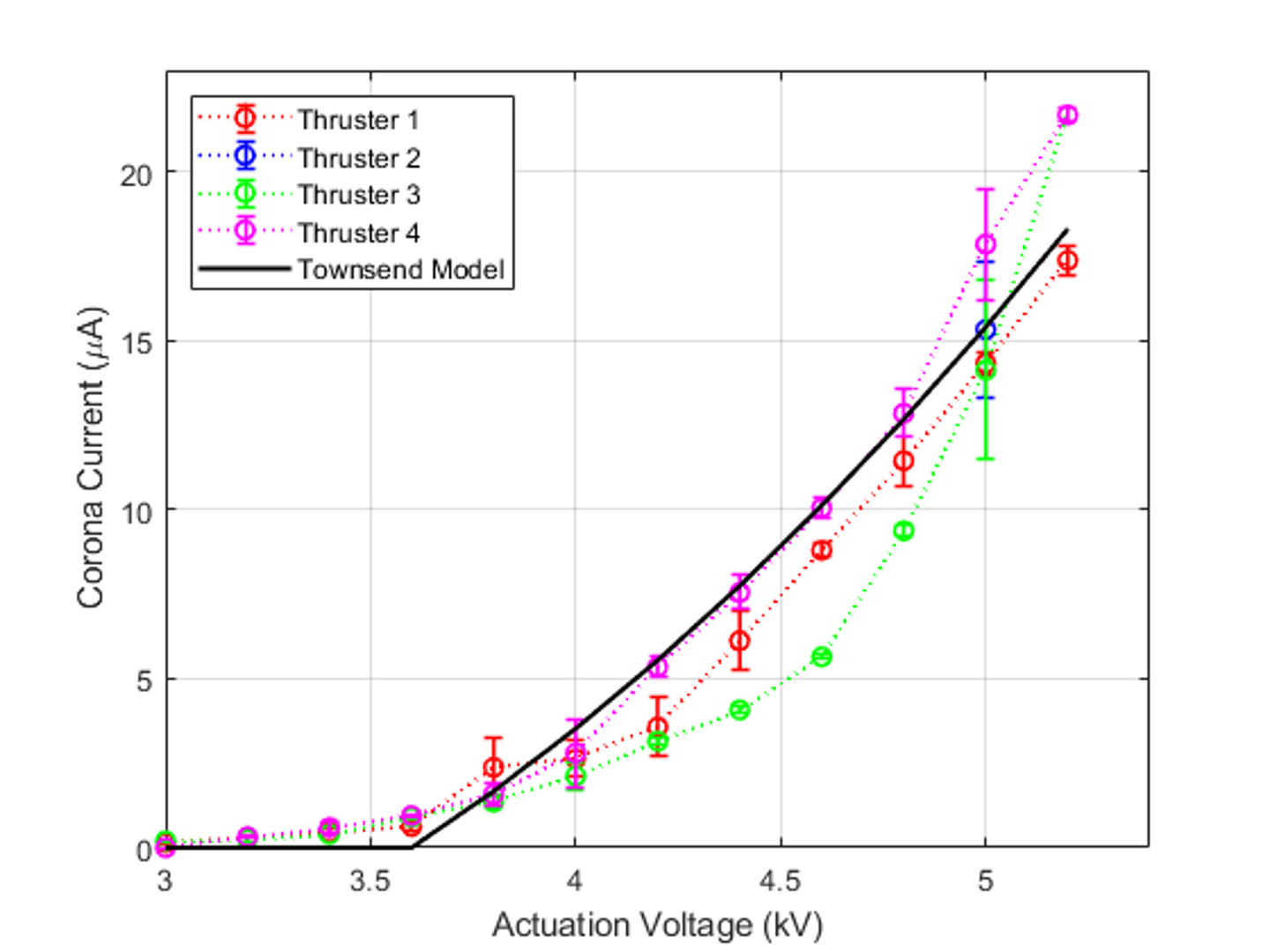}}
\caption{\textbf{Corona current vs the applied voltage for each individual thruster.} The measurements are fitted with a Townsend current model and accurately capture the theoretical trend. Calculated corona onset voltage is 3600~V with a standard deviation of 100~V}
\label{fig5}
\end{figure}

Electrical characterization of the individual thruster in the quad thruster was performed first. A high voltage positive DC power supply (Bertan 205B-20R) was used to create the potential difference between the emitter and collector. The current associated with the discharge was determined from the power supply's built-in ammeter. Results are shown in Fig.~\ref{fig5} for each of the four thrusters that comprise a single unit. The current and voltage trends are similar to previously reported quadratic trends \cite{guan2018analytical, townsend1915electricity, townsend1914xi}. Using equation~(\ref{current_eq_townsend}) the current-voltage relationship is plotted, showing that our data agree with the model. A peak force of 1.2 mN for the quad thruster is predicted from Eq.~(\ref{force_eq_townsend}), which corresponds to a thrust-to-weight ratio of 3.5.

\begin{figure}[!htbp]
\centerline{\includegraphics[width=8cm]{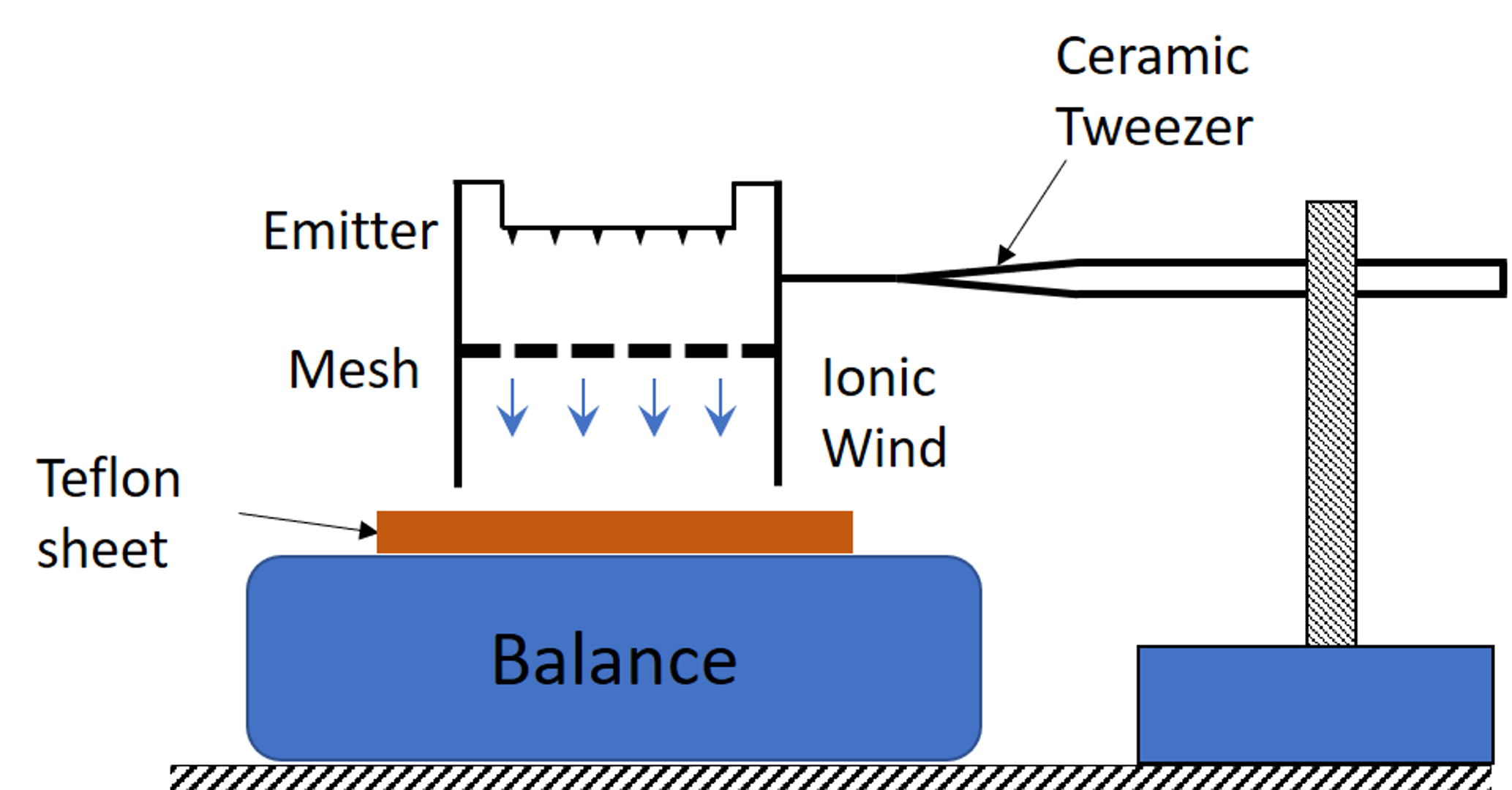}}
\caption{\textbf{Schematic of thrust measurement apparatus.}
Thrust generated by the EHD thruster was measured by measuring the force produced by the ionic wind on the precision scale. Tethers are not shown for simplicity and the robot is not resting on the scale.}
\label{fig6}
\end{figure}

\begin{figure}[!htbp]
\centering
\includegraphics[width=10cm]{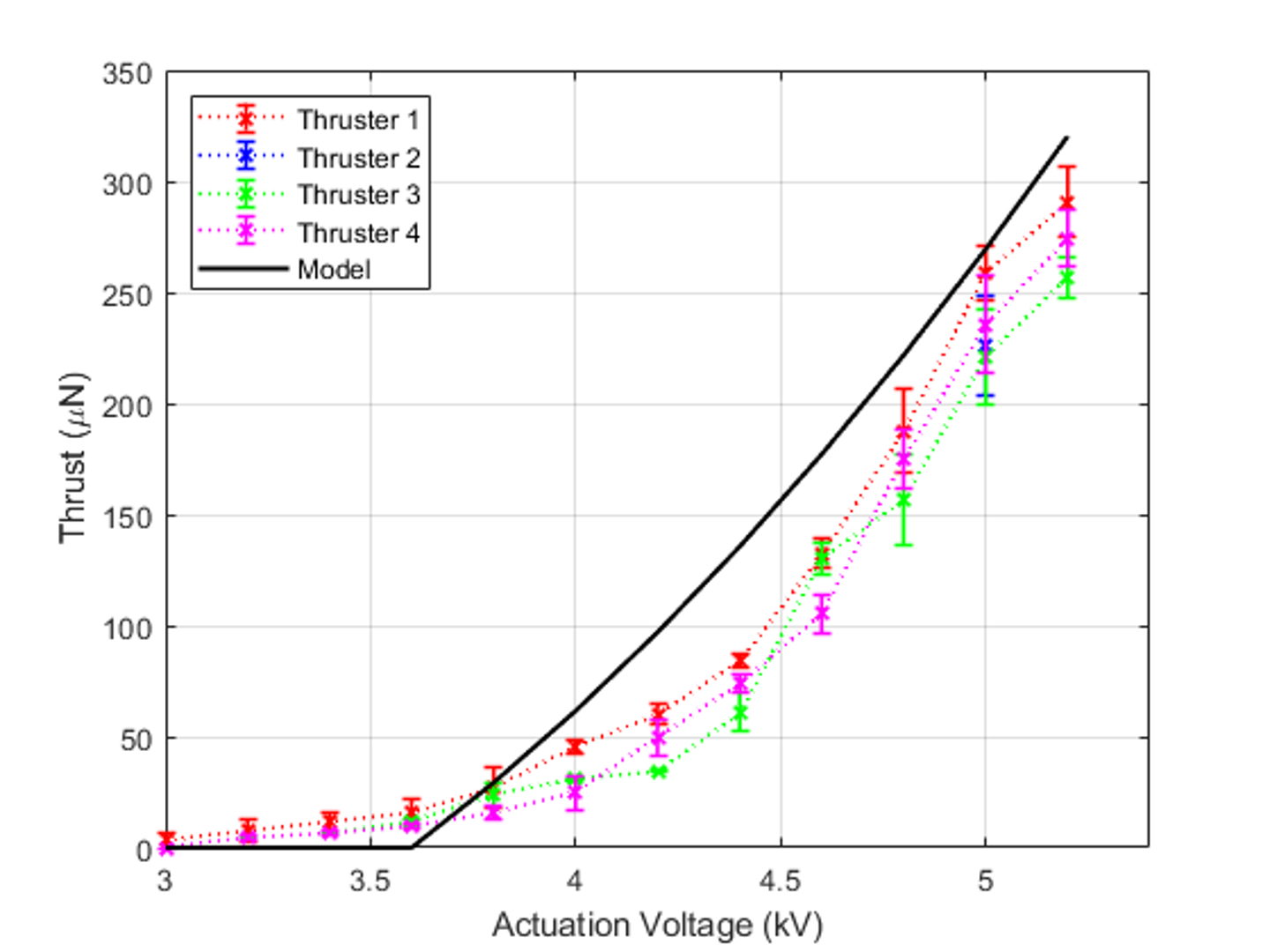}
\caption{\textbf{Thrust variation with applied voltage is depicted for each thruster of the quad-thruster robot.} The input voltages range from 3 kV to 5.2 kV. The Townsend current model shown in fig.\ref{fig6} is used to compute theoretical upper limit for the measured Coulomb force and the data captures this trend.}
\label{fig7}
\end{figure}

The thrust force calculated from the model is validated using experimental thrust measurement as shown in Fig.~\ref{fig6}; A similar setup has been used in previous studies~\cite{moreau2013electrohydrodynamic}. The thruster was held directly above the balance with 0.1 mg resolution (Mettler Toledo) such that the collector grid was aligned parallel to the scale surface using a ceramic tweezer. This arrangement, with the thruster fixed, reduces the confounding effect of electrostatic forces acting on the aircraft through the tether wire. The distance between the collector grid and the weight scale is 21 mm. The scale reading was set to zero and each individual thruster was energized to measure the thrust. A piece of Teflon was placed between the balance plateau and the collector to electrically isolate the balance and to avoid any leakage current. The measured thrust is the force exerted by the accelerated ionic wind on the precision scale. It can be seen that the thrust increases with the voltage applied across the electrodes as shown in Fig.~\ref{fig7}. The thrust trends follow the previously reported quadratic relationship with the applied voltage in~\cite{moreau2013electrohydrodynamic} and~\cite{masuyama2013performance}. The maximum thrust generated occurs immediately before the sparkover is initiated, at which point thrust drops to zero and destroying the mesh. The peak force generated by each thruster was around 260~$\upmu \text{N}$ with a standard deviation of 20~$\upmu \text{N}$. The model predictions were higher than the experimental results, but the model captured the trend. The difference between the model and experiments can be attributed to blockage and drag losses as shown in~\cite{moreau2013electrohydrodynamic}. The model described in Eq.~(\ref{force_eq_townsend}) helped us to validate the thrust measurements and can form the basis for future constrained parametric optimization studies of the quad-thruster robot.

\begin{figure}[!htbp]
\centering
\includegraphics[width=10cm]{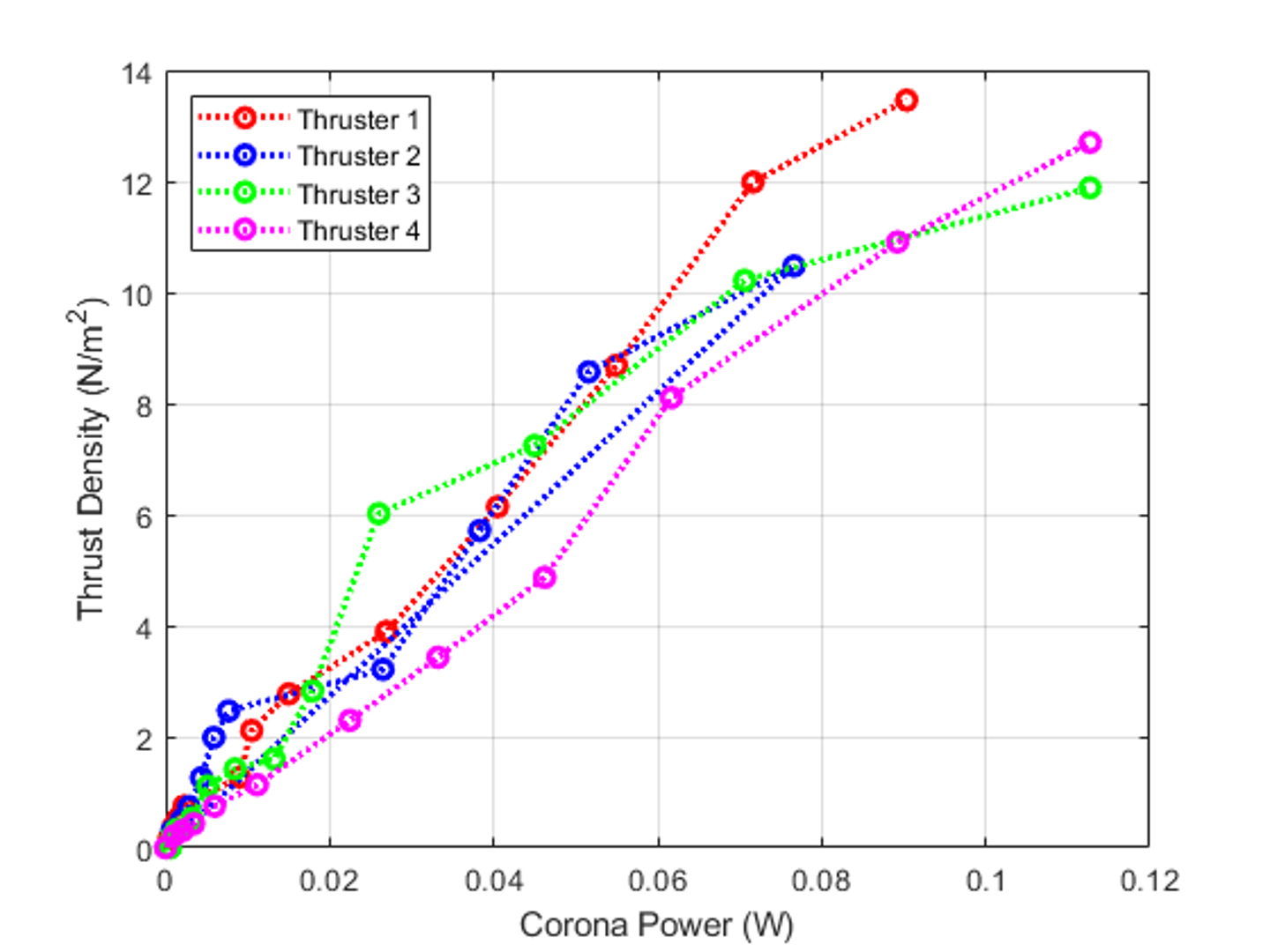}
\caption{\textbf{Efficiency in terms of thrust density versus the corona power for each thruster in the quad-thruster robot.} The data points shown displayed are mean values of thrust density and corona power.}
\label{fig8}
\end{figure}

Thrust density and efficiency are important parameters in understanding the working and performance of EHD thruster compared to other designs. Thrust density is defined as the amount of thrust generated per unit area, whereas efficiency is defined as thrust per unit power. Thrust density for EHD thrusters is calculated from the effective area where EHD flow exists, i.e., the mesh area. Fig.~\ref{fig8} shows how the thrust density varies with the corona input power of EHD thruster. The electrical power was calculated from equation \ref{power_eqn}. A maximum thrust of 0.295~mN corresponds to 13.67~N/\text{m$^2$} thrust density achieved at an input electrical power (aka corona power) of 90.4~mW. Therefore, the thrust density per unit power for the EHD thruster is 151.17~N/\text{m$^2$}W. The efficiency is about 3.265~mN/W. This data can be compared to a piezo-actuated flapping wing such as the RoboFly \cite{chukewad2018}, which has a measured efficiency of 12.2~mN/W. For a thrust of 0.736~mN, input power of 60~mW, and 308~\text{mm$^2$} effective swept area of the wing, the thrust density is 2.39~N/\text{m$^2$}. Therefore, the thrust density per unit power is 39.8~N/{m$^2$W}. Therefore, while the efficiency of the EHD thruster is lower than a flapping-wing robot of comparable size (the efficiency of the flapping wing robot is 3.74 times higher), the thrust density per unit power consumed is 3.8 times higher for the EHD thruster. This is important because the thrust density correlates to the mass of the thruster, and therefore this metric represents a scale-independent (and propulsion-type-independent) measure of efficiency.

\begin{figure}[!htbp]
\centering
\includegraphics[width=13cm]{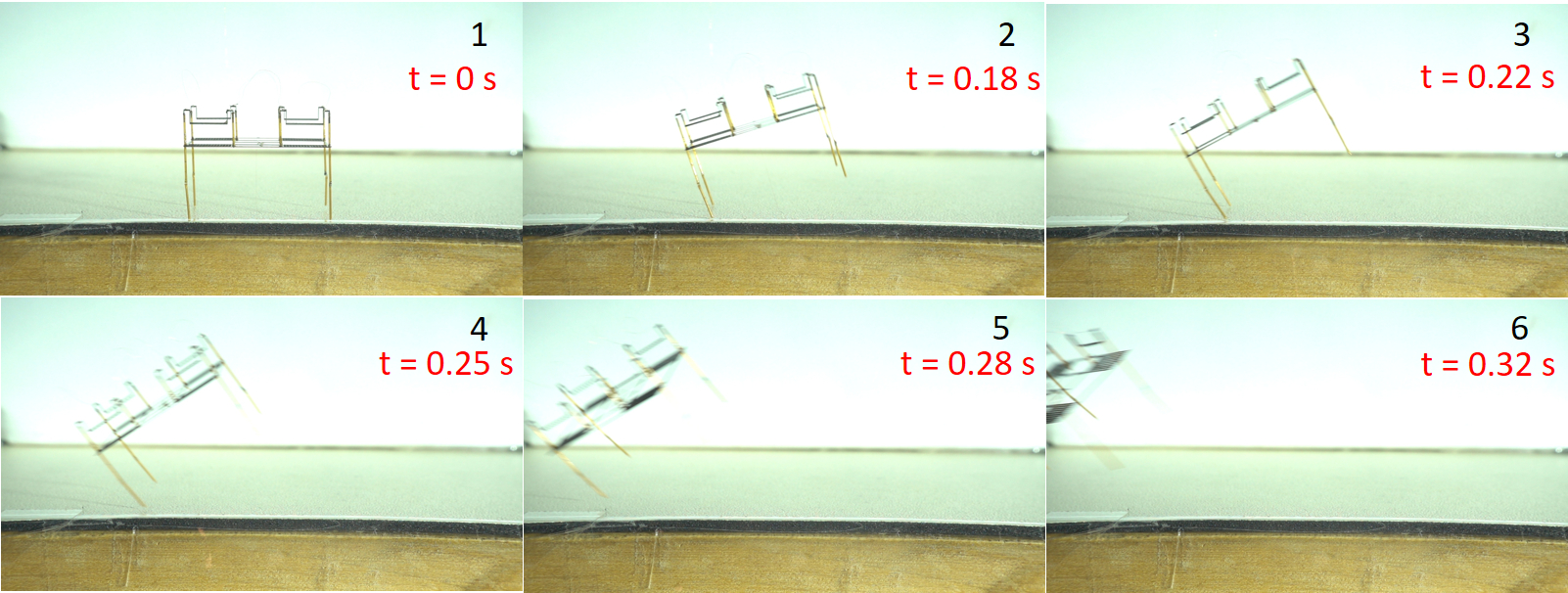}
\caption{\textbf{Frames captured at a frame rate of 240 fps from the quad-thruster in flight.} Robot is resting on the table with the collector connection dangling down and the emitter connection straight up.}
\label{fig9}
\end{figure}

For free flight experiments, the robot was placed on a wooden table. The four emitters are interconnected with 51 gauge wires as mentioned above in the assembly section and the quad thruster was actuated using two 58 gauge copper wires in front of a high-speed camera (Sony RX100). One 58-gauge connection is attached from the top to one of the inner ends of an emitter and the other connection is attached to the center of the collector grid. The power tethers were held using ceramic tweezers and strain relieved. The inner legs were removed and attached to the outer legs of the robot to increase the height and diminish the electrostatic interaction with the takeoff plane. With a voltage of 4.6~kV, lift off of quad thruster was achieved. Fig.\ref{fig9} an image sequence from the flight for the first 0.32 seconds before the wires touched each other, which ended the free flight. This shows conclusively that the robot is able to lift its own weight and we believe that a vertical liftoff can be achieved in future experiments by trimming the robot as illustrated in \cite{DakshDevice}.

Finally, we conclude this section by comparing the performance of the single-emitter thrusters of our four-thruster device to the single-emitter, four-thruster device presented in~\cite{drew2017first} (which did not directly measure thrust measurements). Hence, to compare the performance of the two robots, the thrust generated (estimated from the takeoff video presented in~\cite{drew2017first}) and thrust-to-weight ratio are chosen as the criteria. The single emitter version of the Ionocraft takes off at 2400~V with a corona current measuring close to 20~$\upmu$A. This works out to a corona power of 0.048~W, and from the observed peak acceleration of the Ionocraft, they conclude that the corresponding thrust is approximately 200 $\upmu$N which amounts to a thrust to weight ratio of 2.04. At the same input power, our quad-thruster device generates 675~$\upmu$N which gives a thrust-to-weight ratio of 1.86. Table.\ref{table1} includes a comprehensive summary of manufacturing methodology and performance comparison between the Ionocraft and our laser-fabricated quad-thruster robot.

\begin{table}[!htbp] 
\begin{adjustwidth}{-2.25in}{0in} 
\RaggedLeft
\captionsetup{justification=RaggedLeft} 
\caption{
{\bf Comparison of our work with earlier work by Drew \cite{drew2017first}.}}
\begin{tabular}{|l+l|l|l|l|l|l|l|}
\hline
\multicolumn{1}{|l|}{\bf Comparison Criterion} & \multicolumn{1}{|l|}{\bf Drew \cite{drew2017first}} & \multicolumn{1}{|l|}{\bf Our work}\\ \thickhline
Electrode Material & Silicon & Carbon fiber (grid) \\
\ &  &Stainless steel (emitter) \\ \hline
Total weight ($\upmu$N) & 98~$\upmu$\text{N} (10~\text{mg}) & 362.6~$\upmu$ \text{N} (37~\text{mg}) \\ \hline
Thrust at 0.048~W ($\upmu$\text{N}) & 200~$\upmu$\text{N} & 675~$\upmu$\text{N} \\ \hline
Thrust to Weight ratio at 0.048~W & 2.04 & 1.86 \\ \hline
Assembly time in minutes & 30 & 15 \\ \hline
Clean-room facility & Required & Not required \\ \hline
\end{tabular}
\label{table1}
\end{adjustwidth}
\end{table}

\section*{Conclusion}

A quad thruster was fabricated using UV laser micro-machining  and the performance of the quad thruster is characterized. Measured current and thrust values are in good agreement with the EHD theory. The variation across all the four thrusters in the quad is minimal and the lift off of quad-thruster robot is demonstrated. The quad-thruster robot was able to lift its own weight, as indicated by thrust measurements and free-flight connected to a wire tether. The thrust-to-weight ratio of our robot at takeoff voltage of 4.6~kV is 1.38 with a power consumption of 0.037~W. The peak thrust-to-weight ratio of our robot at the maximum actuation voltage of 5.2kV (with typical operation range of 3.6kV to 5.2kV) is 3.03 with a power consumption of 0.104~W, which is slightly below the ratio of 4.5 reported previously~\cite{drew_hilton_head}. We used fewer emitter rows than~\cite{drew_hilton_head}, and we expect that adding rows of tips, will substantially increase lift with little added weight. We also believe there is ample opportunity to reduce the mass of our device through the use of thinner and lighter material. We plan to conduct an exploration of different designs in finite element simulation to explore the configuration space in greater detail. All of this work will contribute significantly in boosting the thrust to weight ratio of our quad-thruster robot which is fundamental for steps towards autonomy.

The fabrication time, from raw materials to complete assembly takes less than 25 minutes and this approach allows for greater flexibility in the selection of materials. The process is viable and a faster alternative to a silicon-on-insulator fabrication process of an EHD thruster at the laboratory prototyping stage. This compares favorably with the process reported in literature~\cite{drew2017first}, which takes 2-3 days at best. The fabrication time includes next-day shipping time for the masks and other queue time. In addition to short fabrication time, we remark that laser micro-fabrication allows for a much more diverse material set. While silicon has a high strength-to-weight ratio, there are other materials that can provide better performance for certain applications, such as the even higher strength-to-weight ratio of unidirectional carbon fiber composites. Furthermore, if there are other materials that may improve the lifetime of the sharp emitter tips, it is almost certainly possible to machine it using a DPSS laser, and the fabrication of these electrodes does not require a cleanroom facility. 

The path to an autonomy mandates advancement on a number of fronts. Various components are needed as payload. A small on-board camera (as demonstrated in \cite{balasubramanian2018insect} for a flapping wing insect-scale robot) along with other sensing units such as IMUs (has mass of 37~mg in \cite{drew2018}) can be added, which will be instrumental in controlled flight and sensing around surroundings. Future work also includes work on on-board power supply and associated power-electronics. Until recently, micro-robots have been powered through external connections. The Autonomous Insect Robotics Lab at the University of Washington has developed light weight circuit that requires no battery and provides wireless power to a robot \cite{james2018liftoff}. This circuit was capable of developing 200~V to drive piezo actuators in a 100~mg package. We expect that a similar approach could extend to the kV potential differences needed for EHD thrusters.

\section*{Acknowledgment}
The authors also wish to thank Johannes James and Vikram Iyer for insightful discussion during experimental set up and initial design, and TenCate, Inc. for donating the composite materials used in this study.

\bibliography{plos_one_EHD_main.bib}

\pagebreak

\appendix
\section*{Appendix}
\subsection*{Future flight control}

\begin{figure}[!htbp]
\includegraphics[width=10cm]{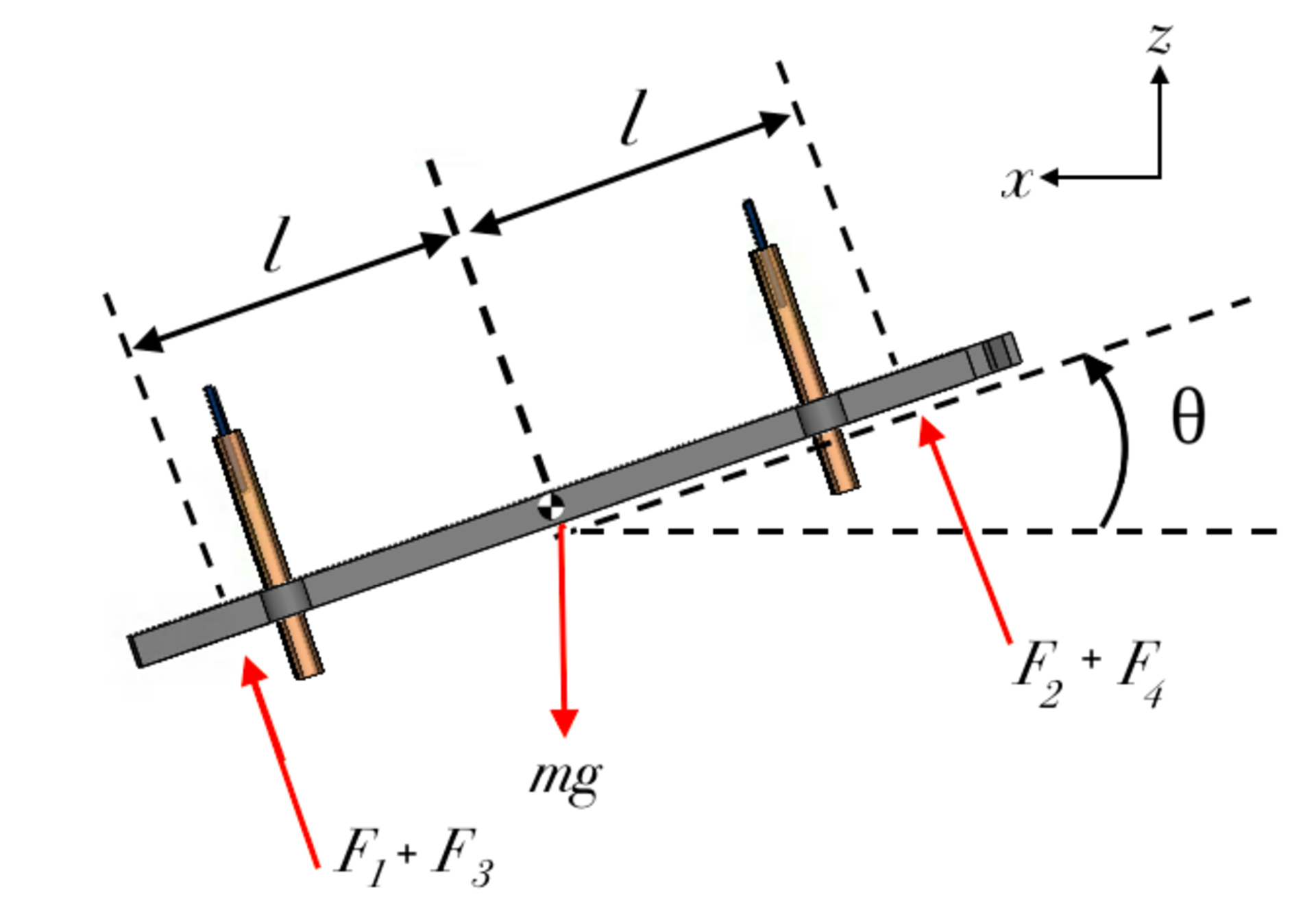}
\caption{\textbf{Free-body diagram of a quad-thruster in a side view to demonstrate the pitch/roll control.} The pitch/roll control of a quad thruster is similar to multicopters. Thrust generated by i-th thruster is denoted by $F_{i}$. For positive pitch/roll angle, $\theta$. $F_{1}$+$F_{3}$ and $F_{2}$+$F_{4}$ can be controlled actively to stabilize the robot at a desired $\theta$.}
\label{fig10}
\end{figure}

The goal of using four individual EHD thrusters is to allow for pitch and roll control similar to that of a quadcopter. We anticipate the controller for the quad-thruster will be similar to that of the four wing insect scale flapping wing robot developed in \cite{fuller2019four} and quadcopters.  The free body diagram for the pitch/roll dynamics is mentioned is shown in Fig.~\ref{fig10}. The robot is symmetric about roll and pitch axis, same formulation can be applied in both directions. Controlling the yaw motion is left for future work. Pitch/Roll dynamics is described as follows.

\begin{equation}
\begin{split}
I_{p}\ddot{\theta} & = \lbrack(F_{2}+F_{4}) - (F_{1}+F_{3})\rbrack l\\
m\ddot{z} & = \lbrack (F_{2}+F_{4}) + (F_{1}+F_{3})\rbrack \cos{\theta} - mg\\
m\ddot{x} & = \lbrack (F_{2}+F_{4}) + (F_{1}+F_{3})\rbrack \sin{\theta}\\
\end{split}
\end{equation}
Derived from Fig.~\ref{fig10}, where $I_{p}$ is the moment of inertia of the robot about the pitch/roll axis, $\ddot{\theta}$ is the angular acceleration, m is the mass of the robot, and $F_{i}$ is the thrust force generated by i-th thruster. This shows that by varying thruster forces $F_i$, the robot is fully controllable in the $x-z$ plane.

\end{document}